\documentclass{article}

\newif\ifpreprint
\preprinttrue         

\PassOptionsToPackage{numbers,sort&compress}{natbib}

\ifpreprint
  \usepackage[preprint]{neurips_2025}  
\else
  \usepackage[final]{neurips_2025}     
\fi

\usepackage[utf8]{inputenc} 
\usepackage[T1]{fontenc}    

\usepackage{graphicx}
\usepackage{amsmath,amssymb}   
\usepackage{amsfonts}          
\usepackage{xcolor}
\usepackage{subcaption}
\usepackage{rotating}
\usepackage{framed}
\usepackage{microtype}         
\usepackage{nicefrac}          
\usepackage{url}               
\usepackage{multicol}          
\usepackage[normalem]{ulem}    
\usepackage{scalefnt}          

\usepackage{hyperref}
\hypersetup{
  colorlinks=true,
  linkcolor=black,
  citecolor=black,
  urlcolor=black,
  pdfauthor={},
  pdftitle={}
}
\usepackage[nameinlink,noabbrev]{cleveref}

\usepackage{booktabs}
\usepackage{multirow}
\usepackage{tabularx}
\usepackage{makecell}
\usepackage{array}
\usepackage{ragged2e}

\newcolumntype{Z}{>{\raggedleft\arraybackslash}p{30mm}}
\newcolumntype{Y}{>{\raggedleft\arraybackslash}X}

\renewcommand{\arraystretch}{1.1}
\usepackage{algorithm}
\usepackage{algorithmic}

\usepackage{pifont}        
\usepackage{adjustbox}     
\usepackage{comment}

\usepackage{fancyhdr}
\makeatletter
\providecommand{\@noticestring}{}
\renewcommand{\@noticestring}{}  
\makeatother
\fancypagestyle{firstpage}{\fancyhf{}\fancyfoot[C]{\thepage}}

\providecommand{\bmhead}[1]{\vspace{1ex}\noindent\textbf{#1}\quad}

\title{Illusions of reflection: open-ended task reveals systematic failures in Large Language Models' reflective reasoning}

\author{%
  Sion Weatherhead\textsuperscript{*} \\
  University of New South Wales\\ Sydney, Australia \\
  \texttt{s.weatherhead@unsw.edu.au}
  \And
  Flora Salim\textsuperscript{*} \\
  University of New South Wales\\ Sydney, Australia \\
  \texttt{flora.salim@unsw.edu.au}
  \And
  Aaron Belbasis \\
  Aurecon Group \\
  Melbourne, Australia\\
  \texttt{aaron.belbasis@aurecongroup.com.au}
}

\begin{document}

\maketitle
\begin{center}
\vspace{0.5em}
\textsuperscript{*}\textit{Corresponding authors.}
\end{center}
\begin{abstract}
Humans do not just find mistakes after the fact—we often catch them mid-stream because `reflection’ is tied to the goal and its constraints. Today’s large language models produce reasoning tokens and `reflective’ text, but is it functionally equivalent with human reflective reasoning? Prior work on closed-ended tasks—with clear, external `correctness’ signals—can make ‘reflection’ look effective while masking limits in self-correction. We therefore test eight frontier models on a simple, real-world task that is open-ended yet rule-constrained, with auditable success criteria: to produce valid scientific test items, then revise after considering their own critique. First-pass performance is poor (often zero valid items; mean $\approx$ 1), and reflection yields only modest gains. Crucially, the second attempt frequently repeats the same violation, indicating `corrective gains’ arise largely from chance production of a valid item rather than error detection and principled, constraint-sensitive repair. Performance before and after reflection deteriorates as open-endedness increases, and models marketed for `reasoning’ show no advantage. Our results suggest that current LLM `reflection’ lacks functional evidence of the active, goal-driven monitoring that helps humans respect constraints even on a first pass. Until such mechanisms are instantiated in the model itself, reliable performance requires external structure that enforces constraints. Our code is available at: https://github.com/cruiseresearchgroup/LLM\_ReflectionTest
\end{abstract}

\section*{Introduction}
Large language models (LLMs) are increasingly entrusted with planning, drafting, and analysis. Step-by-step prompting `chain-of-thought') can boost performance\cite{wei2022ChainofThoughtPromptingElicits}, and recent work encourages models to `reflect' via self-critiques, reflection loops (routing outputs back into a new attempt), and tool-augmented checks\cite{shinn2023ReflexionLanguageAgents,renze2024SelfReflectionLLMAgents,dhuliawala2024ChainofVerificationReducesHallucination}. We use \textit{reflection} as an umbrella for these procedures—including iterative refinement\cite{madaanSELFREFINEIterativeRefinement,tong2024CanLLMsLearn}, code- or test-based feedback\cite{chen2023TeachingLargeLanguage,dhuliawala2024ChainofVerificationReducesHallucination, gou2024CRITICLargeLanguage}, and program-level scaffolding for long-reasoning models (LRMs)\cite{chen2025ReasoningEraSurvey}. 

These effort aim to \emph{functionally} approximate human \textit{meta-reasoning} to improve LLM reasoning. Meta-reasoning refers to a goal-directed process of monitoring and controlling one's thoughts, thought processes(i.e., reasoning), while calibrating efforts and strategy in accordance with confidence, updating allocation of effort and resources whilst maintaining constraint fidelity\cite{ackerman2017MetaReasoningMonitoringControl}. This is a fundamental mechanism for \emph{self-correction}. However, to note, our standard for evaluation is explicitly \emph{functional}, not mechanistic: we do not require human-like processes inside the model, only that the model’s outputs reliably indicate constraint-sensitive evaluation and corrective change.

Why does this matter for intelligence? Broadly, intelligence is the capacity to learn, adapt, and solve problems across familiar and novel settings\cite{sternberg2011CambridgeHandbookIntelligence}. Meta-reasoning is a core human capability for doing so: it detects violations, adjusts strategy, and prevents error repetition\cite{ackerman2017MetaReasoningMonitoringControl}. If LLMs are to be genuinely self-improving rather than merely verbose, `reflection' must act as such a controller. Surveys and targeted studies, however, suggest that many gains attributed to reflection depend on \emph{external} signals—tests, retrieval, or explicit error flags—with limited success at autonomous corrective reasoning when those signals are weak\cite{renze2024SelfReflectionLLMAgents,kamoi2024WhenCanLLMs,huang2024LargeLanguageModels,hosseini2024NotAllLLM}.

A second, deeper concern is the reliability of the `reasoning' text itself. Performance can degrade under superficial changes that leave underlying logic intact—for example, swapping surface values in GSM-style maths questions\cite{mirzadeh2024GSMSymbolicUnderstandingLimitations} or reframing otherwise identical problems\cite{heo2025LLMsknowinternally,jiang2024PeekTokenBias}. More critically, explanatory tokens need not reflect the latent decision process: models may answer correctly while their explanations omit, misattribute, or contradict the basis for the answer\cite{StopAnthropomorphizingIntermediate,chenReasoningModelsDont}. Recent work even shows scale-related failures when task complexity increases while problem structure is held constant\cite{shojaee2025IllusionThinkingUnderstanding}. If traces are decoupled from mechanism, then longer traces and reflection loops risk compounding confident text rather than delivering targeted correction.

These points motivate a distinction that matters for evaluation. \emph{Closed-ended} tasks have a single correct answer and offer crisp corrective signals (e.g., a failing unit test). Many positive reports of reflection occur in such settings\cite{renze2024SelfReflectionLLMAgents}. \emph{Open-ended} tasks enlarge the solution space and require consistent application of interacting constraints; external signals are weaker or delayed, and there are many more plausible but wrong outputs. If LLM behaviour is, in part, statistical pattern matching\cite{mirzadeh2024GSMSymbolicUnderstandingLimitations}, open-ended tasks remove the crutch of strong signals, making principled, constraint-sensitive reasoning harder. We found two examples of open-ended task studies; one was an abstract writing task - where they found systematic biases (e.g., over-generalising study findings) resistant to prompted correction\cite{Generalizationbiaslarge}. The other was a creativity task, where evaluation was either subjective human judgement or relying on dispersion metrics that, while concrete, are hard to interpret as practical \emph{reasoning} diagnostics\cite{gooding2025Writingtestbedopen}.

As such, there is a need for objective, measurable, open-ended tasks for evaluation. Our diagnostic task asks for the minimum behavioural evidence one should expect from a reflective system: \emph{consistency between stated criteria and the change on the next pass}. When a model itself flags plagiarism or a violation of CRT properties, does its revision \emph{avoid that exact fault}? Failure to do so indicates that the \emph{nested concepts} implicated by the task (e.g., what qualifies an `original questionnaire item') are not being activated relative to the stated goal—a shortfall in functional meta-reasoning\cite{ackerman2017MetaReasoningMonitoringControl}. In other words, we seek outputs whose reasoning text \emph{suggests} the constraints have been considered and enforced, without assuming a human-like mechanism.

To probe this directly, we evaluate eight frontier LLMs on an \textit{open-ended}, rule-constrained psychometric generation task with auditable pass/fail criteria. Models must produce Cognitive Reflection Test (CRT) items. These are `trick questions' with an intuitive but wrong answer and a single correct answer reachable upon reflection\cite{frederick2005}. For example:

\begin{quote}\small\itshape
A bat and a ball cost \$1.10 together. The bat costs \$1.00 more than the ball.\\
\textbf{How much does the ball cost?}
\end{quote}

The intuitive response people are drawn toward is `10 cents' from subtracting a \$1 from the initial sum of \$1.10. Upon reflection, along with simple maths, the correct answer of 5 cents can be arrived at. Having an `intuitive-incorrect' response, a common logical error, is a key property of a valid CRT item. 

The LLMs in our tasks must produce new items for this test without copying existing items from validated tests in the literature. We compare two framings that vary the task’s openness and the availability of external anchors: \textit{generation} (invent items de novo) versus \textit{search--identify} (retrieve suitable non-CRT trick questions and adapt). We then solicit a reflection pass and a re-answer. To maximise models’ chances, we provide chain-of-thought scaffolding and an expert-persona prompt, set LRMs to high-reasoning modes where available, allow generous thinking budgets, and use independent LLM judges with access to compendia of known CRT items for plagiarism screening\cite{sprague2024CoTnotCoT,renze2024SelfReflectionLLMAgents}. This design operationalises the intelligence-relevant question: does today’s LLM `reflection' reliably convert self-explanation into \emph{correction} when the problem is open-ended, the constraints are explicit, and the change is auditable?

\paragraph{Not a creativity benchmark.}
Despite the open-ended format, this study does \emph{not} judge `creativity'. We do not score `originality' by aesthetic value. Passing requires only auditable constraints: (i) non-plagiarism (binary check: `does item $x$ appear in set $Y$?'); (ii) ensure properties: e.g., `intuitive-incorrect response' and `correct answer' are present; and (iii) basic clarity (further defined below). Otherwise, we are deliberately lenient: items need not be `publishable' CRT questions— the evaluation target is constraint adherence, not `novelty' per se.

\begin{table}[t]
\caption{Design and outcome counts (pooled across models). Unit: items unless noted. Each session attempts 4 items initially; failed items then enter a reflection phase. “Categorised failures (base)” is the denominator for same-category repeats.}
\label{tab:design_counts_compact}

\footnotesize
\renewcommand{\arraystretch}{1.1}
\setlength{\tabcolsep}{6pt}
\begin{tabularx}{\linewidth}{l r r r Y}
\toprule
Task & N\_sessions 
& \makecell[c]{Initial pass\\(x/expected)} 
& \makecell[c]{Categorised failures\\(base)} 
& \multicolumn{1}{c}{\makecell{Same-category repeats\\(\% of base)}} \\
\midrule
Generation       & 32 & 22 / 128 & 68  & \mbox{58 / 68 (85.3\%)} \\
Search\textendash identify & 32 & 37 / 128 & 52  & \mbox{39 / 52 (75.0\%)} \\
\midrule
Total            & 64 & 59 / 256 & 120 & \mbox{97 / 120 (80.8\%)} \\
\bottomrule
\end{tabularx}

\par\noindent
\vspace{3pt}
\raggedright\footnotesize
\textbf{Notes.} `Initial pass' = valid initial items; expected $=4\times N_{\text{sessions}}$. “Categorised failures (base)” = initial invalid items assigned a failure category; this is the denominator for “Same-category repeats”. For completeness, the \emph{reflection-failure} denominator gives a similar picture: overall $484/567$ (=85.4\%) same-category repeats among reflection failures, with per-strategy rates—retry: $127/143$ (88.8\%), instructions: $127/150$ (84.7\%), explanation: $108/128$ (84.4\%), keywords: $122/146$ (83.6\%). Post-reflection performance is reported as paired \emph{pass-rates} in Table~\ref{tab:passrates}; because those rates average across strategies within session, they are not converted to unique item counts here.
\end{table}

\section*{Hypotheses}

We tested three hypotheses (H). `Pass-rate' is the share of valid items per session (4 items/session). 

\textbf{H1 — Reflection improves performance.}
\textit{H1a (overall):} Reflection increases session pass-rate relative to the initial attempt. 
\textit{H1b (error repetition):} Models will repeat the same failure category in reflection, beyond what would be expected by a chance-based benchmark.

\textbf{H2 — Task structure moderates gains and error persistence.}
\textit{H2a:} Gains are larger in \emph{search–identify} than in \emph{generation}. 
\textit{H2b:} Same-category failure repetition is lower in \emph{search–identify} than in \emph{generation}.

\textbf{H3 — `Reasoning-model' advantage.}
Models marketed for extended reasoning achieve larger reflection gains than other models.

Differences across reflection strategies (Explanation, Retry, Keywords, Instructions) are summarised in the main for context and fully tested in the Supplementary with multiplicity control. Benchmark construction, evaluator setup, and robustness checks are detailed in Methods/Supplementary.

\section*{Methods}
We evaluate eight frontier large\-language models (LLMs) on a three-stage \textit{session} that attempts to invent four novel items for the Cognitive Reflection Test (CRT)\cite{thomson2016}.  
Each session comprises (i) an initial answer, (ii) self-reflection, and (iii) a re-answer, mirroring the `Baseline' and `self-reflecting agent' scheme of Renze \& Guven\cite{renze2024SelfReflectionLLMAgents}.  

All experiments were conducted using LLM\_ReflectionTest, a modular prompting and evaluation platform developed for this study. The system allows flexible insertion of role, generation, critique, and reflection prompts, supports parallel evaluation across multiple language models, and automatically logs outputs and metadata into a structured database. This platform implements the full generate–evaluate–reflect loop described below.

\subsection*{Models}
We evaluated eight models: OpenAI (GPT-4.1, o3, o4-mini), Google (Gemini 2.5 Pro-Preview), Anthropic (Claude 3.7 Extended), Meta (Llama-3.3-70B, Llama-4 Maverick), and DeepSeek (Reasoner). Where available for LRMs, `high-reasoning’ modes were enabled.

While each model was evaluated over eight sessions, each session constitutes a full agentic cycle involving initial generation, error detection, 3 distinct reflection strategies created, and finally four targeted re-attempts applying the 3 created strategies and a `retry' where models are asked to simply replace failed items with no additional strategy. 

Thus, each model undergoes $8 \times 4 = 32$ structured reasoning attempts across diverse prompt framings (see `task definition' below).

All LLMs were evaluated using their standard public API configurations, including default temperature settings (typically in the 0.7–1.0 range). This choice reflects real-world usage patterns for generative tasks, where deterministic decoding is rare. Crucially, our goal was not to constrain models into a narrow reasoning trace (which could artificially dampen performance on an open-ended task), but to simulate adaptive, open-ended generation and evaluate the robustness of reflection under typical sampling variability.

\subsection*{Task definition}
A valid CRT item must  
(i) present a seemingly obvious but wrong answer,  
(ii) permit a single correct answer reachable with reflection, and  
(iii) differ substantively from published CRT items.  

Two high-level task framings were tested:

\begin{itemize}
\item \textbf{Generation} – create four items \textit{de novo}.  
\item \textbf{Search–identify} – retrieve four suitable `trick questions' from public sources, excluding any existing CRT items.
\end{itemize}

Prompt ablations manipulated the presence of exemplars, extra instructional detail, and explicit practical constraints. Materially, these variations do not change the structure of the task. The aim was only to determine whether complexity, specificity, or provision of prohibited items affected the outcome. 

\subsection*{Agent protocol (single LLM session)}
\begin{enumerate}
  \item \textbf{Baseline answer.} Model receives the task prompt and outputs four candidate items.
  \item \textbf{Self-reflection.} For each failed item (detected automatically; rubric below) the same model is asked to \textit{explain the failure} and produce a short corrective advice block (keywords, explanation, step-by-step instructions).
  \item \textbf{Re-answer.} This advice block/reflection text is prepended to the original task prompt reminding of the constraints and the model attempts the task again.  
\end{enumerate}
This three-call cycle constitutes one \textit{session}. No system-level state is carried across sessions.

\paragraph{System persona.}
Following Renze \& Guven \cite{renze2024SelfReflectionLLMAgents}, each session used a constant system prompt establishing an expert persona: `You are a cognitive science expert and psychometrician with experience designing and validating Cognitive Reflection Test (CRT) items.'

We treat persona as a cue that, if effective, should functionally activate associated constraint-relevant knowledge (e.g., canonical CRT set, exclusion checks) and thus increase rule adherence.

\subsection*{Automated evaluation}
Each candidate item was scored by three independent LLM evaluators (GPT-4.1, o4-mini, Gemini 2.5 Pro-Preview) using a fixed rubric and structured output (one API call per item per evaluator). We use a single aggregation rule across criteria: \emph{fail-fast}—if \emph{any} evaluator flags a required criterion as failing, the item fails.

\begin{itemize}
\item \textbf{Validity (1/0):} requires both an intuitive-incorrect response and a single reflection-based correct answer.
\item \textbf{Novelty (1/0.5/0):} evaluators compare against an embedded list of known psychometrically validated CRT items and then check, in an open-world sense, for widely shared non-CRT ``trick questions'' using the evaluator's internal knowledge. \emph{Generation}: 1.0 if neither a CRT nor a common trick; 0.5 for a superficial CRT variation; 0.0 for an exact/near-exact CRT or an existing non-CRT trick. \emph{Search–identify}: 1.0 for a pre-existing non-CRT trick with no CRT overlap; 0.5 for a superficial CRT variation; 0.0 for an exact CRT match. \emph{For pass/fail we collapse to binary: only 1.0 counts as pass; 0.5/0.0 are treated as fail.}
\item \textbf{Clarity (1/0):} linguistic coherence and an unambiguous solution path; the intuitive-incorrect rationale must be logically described.
\item \textbf{Complexity cap (screen; 0–5 source scale):} arithmetic (0–2), algebraic (0–2), spatial (0–1) using a coarse rubric (0 = none, 1 = basic, 2 = advanced). To preserve CRT-like accessibility, items with total complexity $\ge 3$ are labelled non-CRT-like and \emph{fail the screen}. \emph{This cap is binary in analysis.}
\end{itemize}

\paragraph{Pass logic.}
An item is \emph{valid} if all three binary criteria pass under the fail-fast rule (Validity = 1, Clarity = 1, Novelty = 1.0 for the relevant condition) \emph{and} the complexity cap is satisfied (sum $<3$). Any single evaluator’s fail on a criterion, a Novelty of 0.5/0.0, or total complexity $\ge 3$ renders the item invalid.

\textit{Note on scoring.} The tri-level Novelty and the 0–5 complexity rubric are used to support descriptive robustness checks; all confirmatory tests use the binary decision rule above.

\paragraph{Prompts}
Full evaluator rubric, JSON schema, and all prompts (task, reflection, evaluators) are provided in the public code repository (see Data \& Code Availability).

\paragraph{Rationale for LLM non-CRT novelty checks.}
To preserve closed-book evaluation, we use an LLM-only screen to flag `existing' non-CRT trick questions rather than curate an inevitably incomplete non-CRT list. We justify this on two grounds: First, studies show LLM can be effectively utilised for fact-checking\cite{deverna2024Factcheckinginformationlarge}\cite{kuznetsova2023GenerativeAIwe}, indicating reliable open-world detection of widely shared content. Secondly, memorisation increases along with frequency of data appearance in pre-training; data—duplicated sequences are disproportionately internalised and thus detectable \cite{kandpalDeduplicatingTrainingData}. Given popular riddles and viral `trick questions' are precisely such high-duplication artefacts, an instruction-following LLM serves as a pragmatic first-pass detector for non-CRT reuse.

\subsection*{Human validation}
While rubric is largely objective (mathematical complexity, while subjective, is purposely coarse grained to enable most robust criteria checking), we examined LLM evaluation fidelity via human inspection. A stratified \(20\,\%\) sample (strata: model \(\times\) condition \(\times\) reflection) was double-coded by two graduate students, prior to disagreement resolution, inter-rater agreement was $\kappa = 0.55$. However, all disagreements were resolved through discussion with a final set of labels, as it were matter of pointing to missed CRT items, or resolving misreadings of a given question and their answers. Human vs. collective LLM label had an inter-rater agreement of $\kappa = 0.54$. Further details are specified in supplementary materials. 

\section*{Metrics}
To evaluate whether reflection improves LLM performance, we compared valid-item generation before and after reflection across models. The generation of valid items was computed as follows. In the initial round, each LLM in a given experimental condition was tasked with producing four CRT items that met predefined validity criteria. The number of valid items was divided by four to compute a pass-rate (e.g., 2 out of 4 valid items yields a 50\% pass-rate). 

In the reflection round, each of the four reflection strategies was applied to the subset of failed items from initial round of generation. The resulting new items were evaluated for validity, and the post-reflection pass-rate for each strategy was calculated by combining the successful items from the initial round and the successful reflection-attempts, divided by four. This approach follows the method in \cite{renze2024SelfReflectionLLMAgents} and reflects the conceptual structure of the task: a single, continuous attempt by a given LLM in a given condition (which we define as a session) to complete the generative task and correct prior failures. While separate API calls are made for initial generation, reflection strategy generation and finally the strategy execution - this continuity is facilitated by passing through the condition text and exact task wording in each API inside a session.

The formal calculation is shown in Equation~\ref{eq:passrate}, where the subscript \textit{initial} refers to the model's original attempt, and \textit{reflection} refers to its subsequent reflection-based re-attempts.

\begin{equation}
\begin{aligned}
\text{PassRate}_{\text{initial}} &= \frac{\text{Correct}_{\text{initial}}}{\text{Total}_{\text{initial}}} \\
\text{PassRate}_{\text{reflection}} &= \frac{\text{Correct}_{\text{initial}} + \text{Correct}_{\text{reflection}}}{\text{Total}_{\text{initial}}}
\end{aligned}
\label{eq:passrate}
\end{equation}

\section*{Analysis}

\paragraph{H1a — Overall reflection effect.}
For each generation session we paired the initial pass-rate with the mean pass-rate across all four reflection strategies. We fit a linear mixed-effects model with generation round (initial vs.\ reflection) as a fixed effect and random intercepts and slopes by session to estimate the average improvement attributable to reflection while accounting for within-session dependence. As a complementary check, we conducted a paired $t$-test on per-session pass-rates and report the corresponding paired effect size ($d_z$).

\paragraph{H1b — Does reflection repeat the original failure?}
We tested whether reflection repeats the \emph{same} failure category that appeared at the initial attempt. Analytically, we pooled across tasks and models and estimated the probability of repeating the original category with a session-clustered logistic model including strategy as a fixed effect (per-model proportions in Supplementary). Separately, we benchmarked the observed repeat rate against chance—what it would be \emph{if reflection failures were random draws from the observed mix of categories within each task$\times$strategy cell}—using a stratified permutation test (Methods/Supplementary).

\paragraph{H2a — Task moderation of reflection gains.}
To test whether performance gains depend on task structure, we fit mixed-effects models with generation round, task condition (generation vs.\ search–identify), and prompt subcondition (base, complexity, examples, practical) as fixed effects, and a random intercept by session. Model selection (log-likelihood) retained generation round and the required two-way interactions.

\paragraph{H2b — Task effects on error persistence (any-category repeat and plagiarism recidivism).}
We probed whether error persistence differs by task. We modelled the probability that a reflection attempt repeats the session’s initial failure category as a function of task (generation vs.\ search–identify) with session-clustered logistic models controlling for strategy. We contrast tested plagiarism recidivism specifically (conditional on initial plagiarism), again by task.

\paragraph{H3 — Reasoning–model superiority.}
For the reasoning–model contrast we fitted an OLS model of reflection gain with fixed effects for task and subcondition and a binary indicator for model–type (1 = reasoning, 0 = other), clustering standard errors by model identity. The confirmatory one–sided test asked whether reasoning–labelled models achieved larger gains than other models, with CR1 95\% CIs and wild-cluster bootstrap 90\% CIs/$p$. After this test, we ran an exploratory equivalence check (TOST, $\Delta=\pm0.05$ pass–rate units) as a sensitivity analysis.

\paragraph{Exploratory — Strategy differences.}
To compare reflection framings, we modelled strategy as a fixed effect with a random intercept for session, using $\Delta$ pass-rate (recovered valid items out of four) as the outcome. Strategy effects were estimated relative to `explanation', with Holm-adjusted pairwise contrasts and within-session paired effect sizes.

\section*{Results}
\begin{figure}[htbp]
  \centering
  \includegraphics[width=0.85\linewidth]{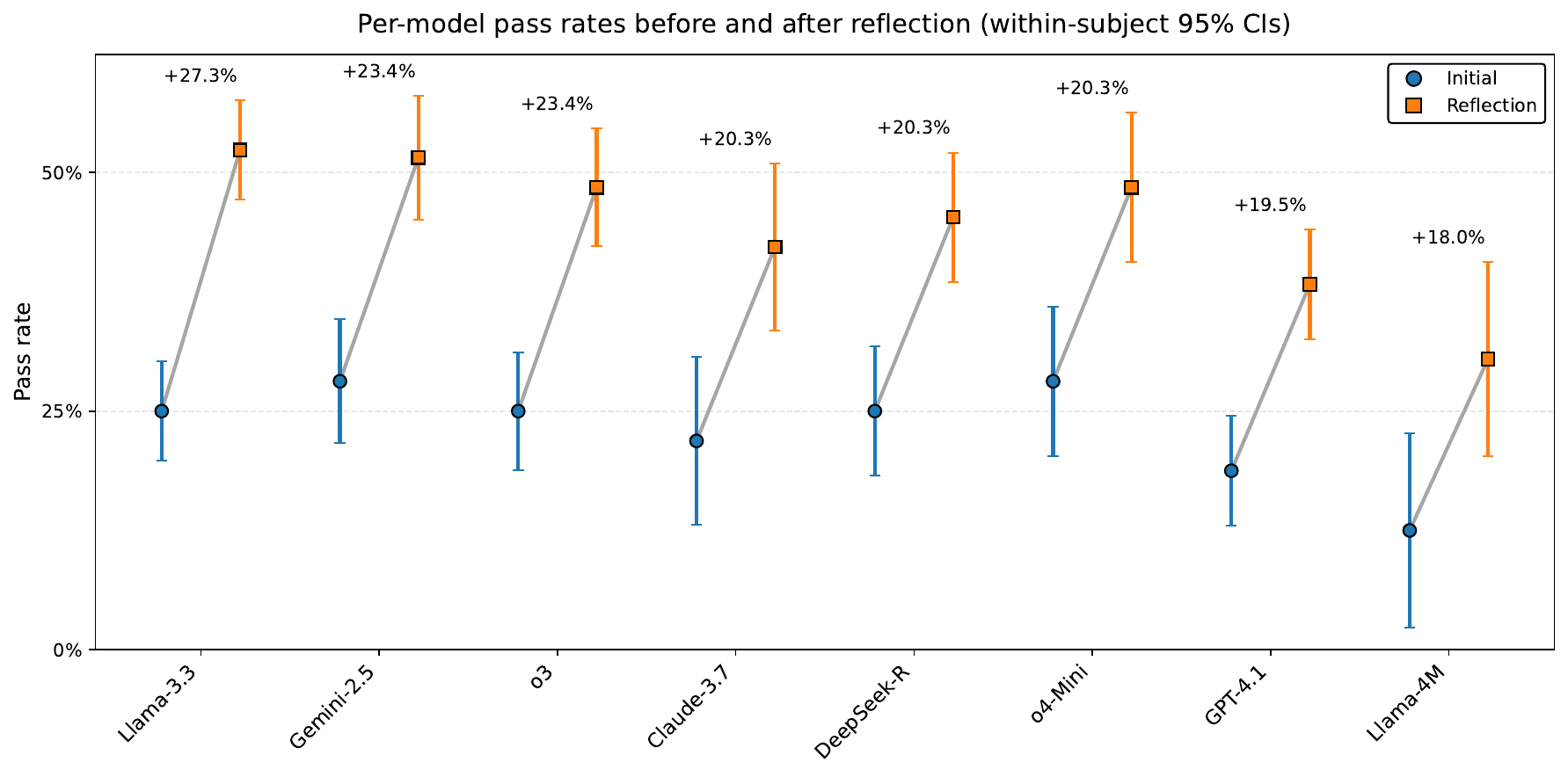}
  \caption{Model-specific gains from reflection. Bars show mean session pass-rates before and after reflection for each model, ordered by the magnitude of improvement. Replicates: sessions; n per model = 8 sessions (4 items/session). Panel shows per-model summaries only; full per-session distributions are in Fig. 2.}
  \label{fig:H4_paired}
\end{figure}

\begin{figure}[htbp]
  \centering
  \includegraphics[width=0.85\linewidth]{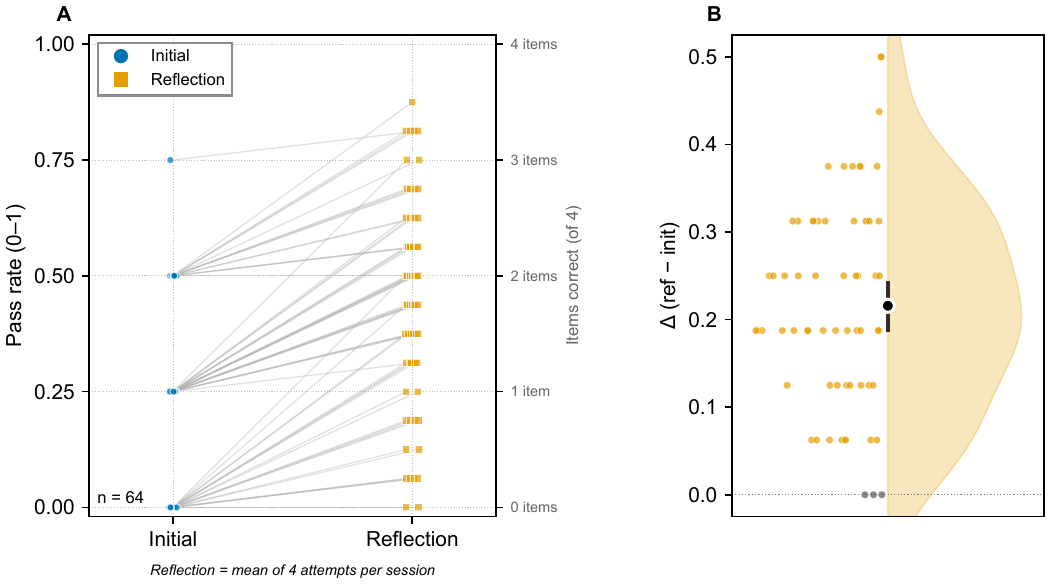}
  \caption{\textbf{H1 — Reflection improves performance.} Panel A Distributions of session pass-rates before and after reflection (n = 64 sessions). Panel B Violin depicts kernel density; central line marks the median; dots show session level variability of delta values; note* reflection pass-rates is produced using the mean across four strategies including `retry'.}
  \label{fig:h1_main}
\end{figure}

\paragraph{H1a — Reflection improves performance.}
Reflection increased pass-rates across all models (Fig.~\ref{fig:h1_main}). A mixed-effects model estimated an average improvement of $\beta = +0.216$, 95\% CI [0.199, 0.232], $p < 0.001$. A paired $t$-test on per-session pass-rates confirmed the effect, $t(63)=14.40$, $p < 0.001$, with a large paired effect size ($d_z = 1.80$).

\begin{table}[t]
\caption{Pass-rate descriptives and paired improvement, \textit{pooled across models and reflection strategies}. Unit: session (4 items/session). `Post-reflection' is the final pass-rate after the reflection phase.}
\label{tab:passrates}
\centering
\begin{tabularx}{\linewidth}{l r c c c}
\toprule
 & N\_sessions & Initial & Post\mbox{-}reflection & $\Delta$ (paired) \\
\midrule
Overall          & 64 & 0.230 & 0.441 & +0.211 \\
Generation       & 32 & 0.172 & 0.281 & +0.109 \\
Search-identify & 32 & 0.289 & 0.602 & +0.313 \\
\bottomrule
\end{tabularx}

\vspace{2pt}
\raggedright\footnotesize
Notes: Initial and post-reflection values are \emph{session means}. Post-reflection counts each item’s final status at the end of the reflection phase (i.e., includes items already correct at initial). Models and reflection strategies are pooled here; strategy effects are analysed separately in the Supplementary.
\end{table}

\paragraph{H1b — Reflection repeats the original failure above chance.}
Across reflection attempts, the repeat-category rate was 85.36\% (567 attempts) when pooling across tasks and models. Relative to a within-cell category-mix benchmark, repetition was higher: observed 85.36\% vs.\ benchmark 74.69\% (95\% permutation interval 72.13–77.25), an excess of +10.68\,pp; permutation $p=0.0001$ (Methods/Supplementary). Strategy terms were not reliably different in the pooled model (all $p \ge .30$).

\begin{figure}[htbp]
  \centering
  \includegraphics[width=0.85\linewidth]{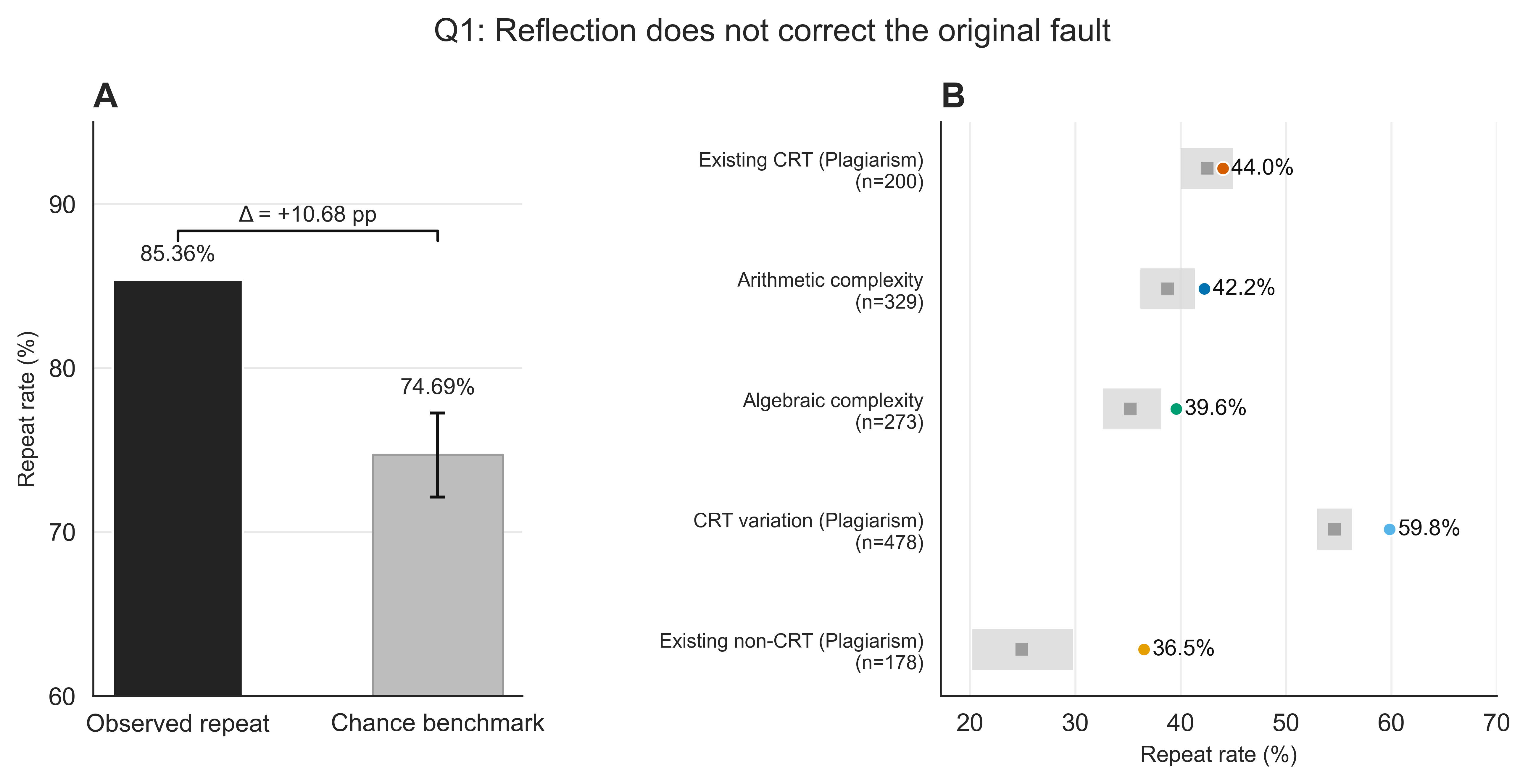}
  \caption{\textbf{H1b — Reflection repeats the original failure above chance.} \textbf{Panel A:} Observed repeat-failure rate at reflection vs.\ a stratified permutation benchmark. \textbf{Panel B:} Five most frequent failure categories. Grey boxes show the 95\% permutation envelopes; coloured dots are observed repeat rates; $n$ per row is the number of reflection attempts at risk (category present initially). Dots right of the envelope indicate above-chance repetition.}
  \label{fig:q1_repeat}
\end{figure}

\begin{figure}[htbp]
  \centering
  \includegraphics[width=0.85\linewidth]{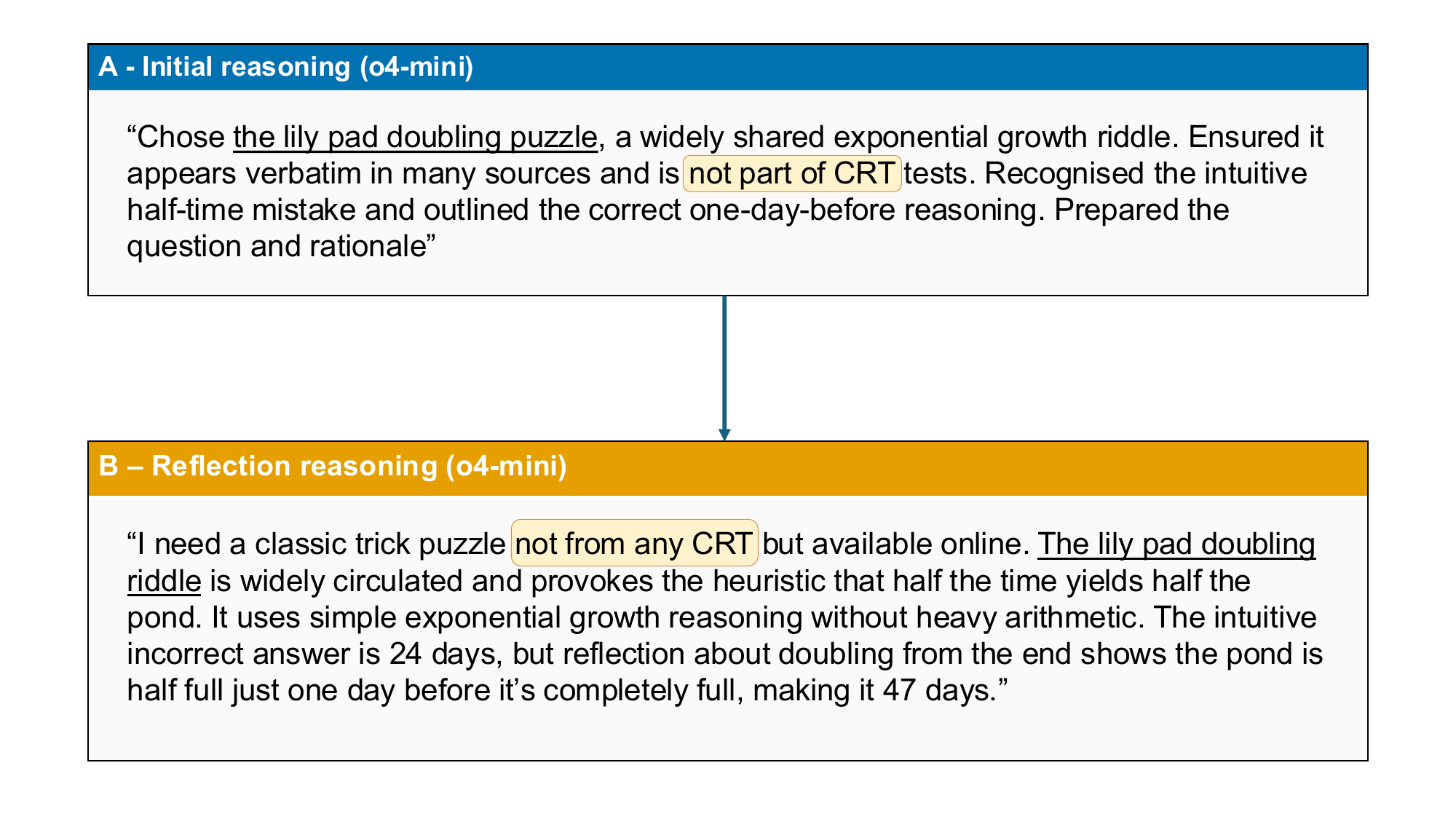}
  \caption{\textbf{Vignette of o4-mini reasoning} Here, o4-mini outputs text reasoning for choosing a CRT item (underlined text) our task prohibits , And fluent mention (highlighted portion) of the constraint not to copy from existing CRT items.}
  \label{fig:vignette}
\end{figure}

\begin{figure}[htbp]
  \centering
  \begin{subfigure}[b]{0.48\linewidth}
    \centering
    \includegraphics[width=\linewidth]{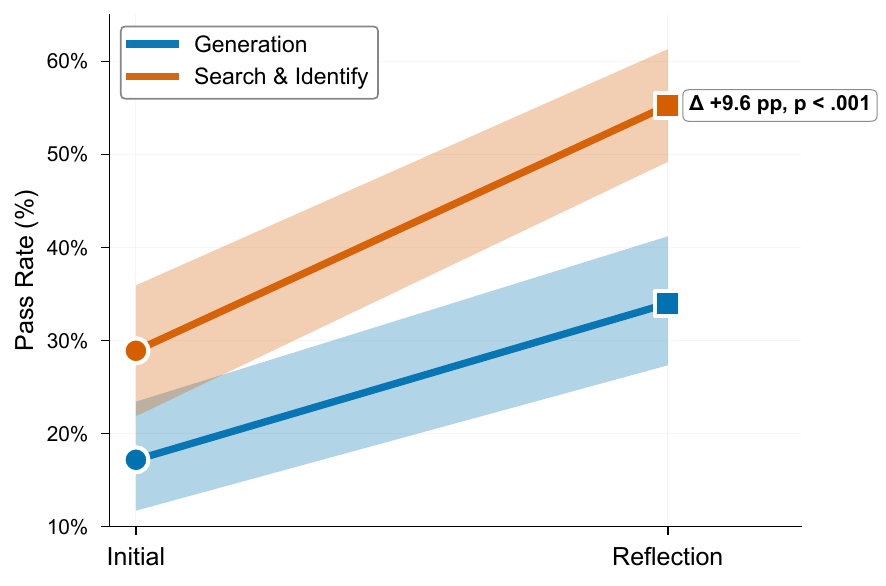}
    \caption{Reflection benefit by task (H2A)}
    \label{fig:h2a}
  \end{subfigure}
  \hfill
  \begin{subfigure}[b]{0.48\linewidth}
    \centering
    \includegraphics[width=\linewidth]{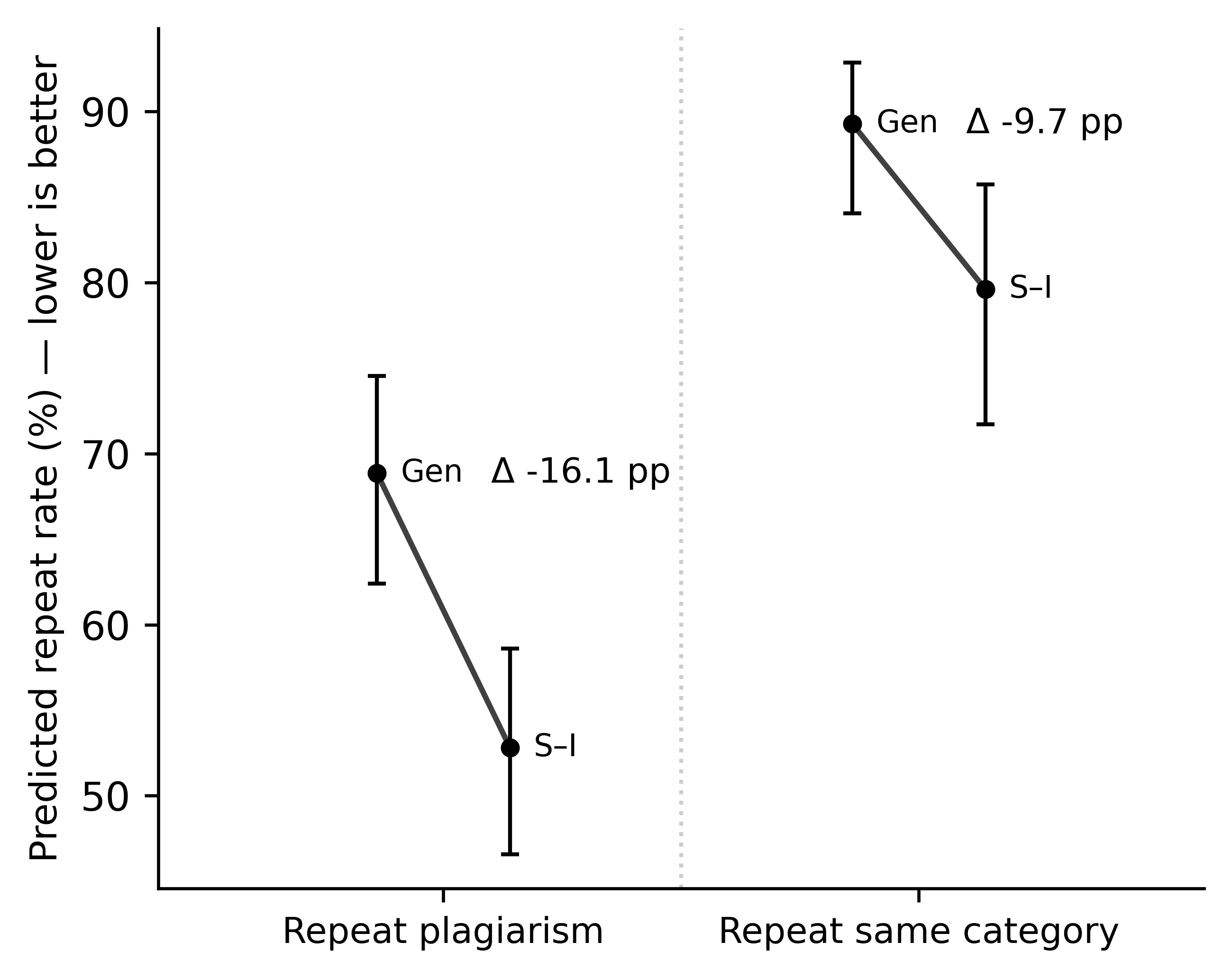}
    \caption{Error persistence by task (H2B)}
    \label{fig:h2b}
  \end{subfigure}
  \caption{\textbf{H2A/H2B — Task structure moderates reflection.} \textbf{(A)} Shows search–identify yields larger reflection gains than generation. \textbf{(B)} Odds of repeating the original category are lower in search–identify; plagiarism recidivism is also lower}
  \label{fig:combined}
\end{figure}

\paragraph{H2a — Task moderates reflection gains.}
Gains were larger for search–identify than generation (Fig.~\ref{fig:h2a}; $\beta = +0.096$, 95\% CI [0.069, 0.122], $p < 0.001$). Among prompt subconditions, only `examples' provided an incremental benefit over the base prompt ($\beta = +0.043$, 95\% CI [0.005, 0.081], $p = 0.025$).

\paragraph{H2b — Task reduces error persistence (any-category repeat and plagiarism).}
Error persistence favoured search–identify (Fig.~\ref{fig:h2b}). Repeating the original category had lower odds in search–identify vs.\ generation (OR $=0.467$, 95\% CI [0.249, 0.877], $p=.018$), controlling for strategy. Plagiarism recidivism—conditional on initial plagiarism—was also lower in search–identify (OR $=0.501$, 95\% CI [0.342, 0.736], $p<.001$).

\paragraph{H3 — Reasoning-model contrast.}
Reasoning–models showed no superiority. Mean reflection gain was 0.036 (SD 0.237, $n=40$) vs.\ 0.111 (SD 0.171, $n=24$) for other models; difference $=-0.075$ pass–rate units ($\approx -0.30$ items out of 4). Model–type coefficient $\beta=-0.075$, CR1 95\% CI $[-0.113,\,-0.037]$, one–sided test $p=0.9999$. Wild cluster bootstrap 90\% CI $[-0.104,\,-0.046]$, one–sided bootstrap $p=1.0$. The exploratory TOST against $\pm0.05$ found the bootstrap CI below the lower bound (suggesting a disadvantage).

\begin{figure}[htbp]
\centering
\includegraphics[width=0.85\textwidth]{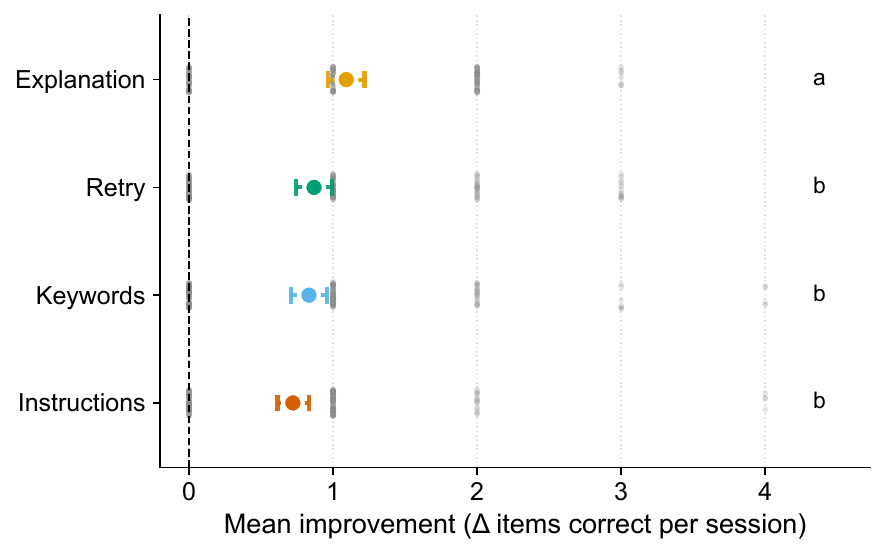}
\caption{\textbf{Exploratory — Effectiveness of reflection strategies.} Mean improvement in valid items per session ($\Delta$ out of 4) with 95\% bootstrapped CIs. Rug plots show per-session values. Compact letter display indicates Holm-adjusted groupings, where contrast results are denoted with \textbf{a} for significance, \textbf{b} for non-significance  ($\alpha=.05$).}
\label{fig:h2_reflection_strategies}
\end{figure}

\paragraph{Exploratory — Strategy differences.}
All four framings improved performance (Fig.~\ref{fig:h2_reflection_strategies}). `Explanation' had the largest mean gain and exceeded other framings after multiplicity control; however, \emph{Retry}—with no explicit reflective scaffolding—was statistically indistinguishable from \emph{Instructions} and \emph{Keywords}, and within–model contrasts for `Explanation' were non–significant in 6/8 models (Supplementary). Practically, the edge of `Explanation' corresponds to roughly one additional item recovered.

\section*{Discussion}
Our experiment indicates that open-ended tasks indeed reduce LLMs ability to complete simple tasks. While reflection helps LLMs in aggregate, the practical effect is far less pronounced than in \cite{renze2024SelfReflectionLLMAgents}'s closed-ended benchmarks. More importantly, LLMs frequently repeat the same mistake committed in the initial round. As seen in Fig.~\ref{fig:h1_main}, gains are small on average and uneven across sessions, while the reflection pass often \emph{repeats} the original failure (85.36\%; Fig.~\ref{fig:q1_repeat}). A system with functional `meta-reasoning' would be expected to convert a second attempt into reliable correction across items, especially for simple yet key errors such as plagiarism -and the problem is even more stark in the Generation condition, where the solution space is larger.

Contrary to expectation, LRMs marketed for extended reasoning showed no reflection advantage over non-reasoning peers after multiplicity control. Our post-hoc equivalence tests suggest a small disadvantage as group. However, our per-model ANOVA showed showed no statistical difference  (see \Cref{ed:model-pass-rates}). In short, longer traces of LRMs combined with our reflection scaffolding did not yield a functional, reliable mechanism that prevents the same rule violation from resurfacing.

The present task type matters because real-world tasks are often open-ended: large solution spaces, weak anchors, and hard constraints. When we enlarged the solution space further (Generation), initial success also drops comparatively, and reflection recovers less than in search-identify (initially 17\% vs.\ 29\%; reflection advantage for S-I: $\beta=+0.096$, 95\% CI $[0.069,\,0.122]$, $p<0.001$; Figs.~\ref{fig:h2a},\ref{fig:h2b}). This pattern suggests that LLM reasoning in both initial and reflection rounds fails to bind to specified constraints. Consistent with this, the combination of high repeat-failure category rates and only modest reflection gains point to a deeper issue with these `gains' made in reflection (see \Cref{tab:design_counts_compact}).

Even when scores improve, the gains are not consistent with principled diagnosis and systematic correction of specific errors. Instead, it appears to be a chance event; resulting from having another attempt and occasionally producing valid items among continued failures. This is not indicative of a mechanism that identifies faults and prevents reoccurrence through principled application of reason. This mechanism is most visible in the results showing error persistence. Across all models, reflection repeats the session's original failure category well above a within-cell chance benchmark (Fig.~\ref{fig:q1_repeat}). At the session level, improvement covaries with how many items enter reflection: when more failed items are retried, the probability of recovering at least one valid item rises and the variance in gains widens—behaviour consistent with second-chance sampling rather than targeted repair (cf.\ Fig.~\ref{fig:h1_main}). Plagiarism is the clearest and most damaging case: after an initial plagiarism flag, many reflections result in further plagiarism—sometimes of the exact same kind (Fig.~\ref{fig:q1_repeat}\textbf{B}; Fig.~\ref{fig:h2b}). 

Our vignette aligns with the statistics: in the search–identify condition, a model explicitly reasons that the lily-pad exponential-growth riddle is `widely shared' and `not part of CRT tests' (incorrect; it is a canonical CRT item), then reproduces that very item; on reattempt, it justifies the same choice and reproduces it again. The reflection text summons the right labels (`do not copy', `not a CRT item') but fails to activate the nested checks that would control generation (`is this in the reference set?', `does this violate novelty?'). The outcome is fluent self-critique without correction. 

Notably, in our open-ended setting, a simple \emph{Retry} was statistically indistinguishable from \emph{Instructions} and \emph{Keywords} after multiplicity control \cref{fig:h2_reflection_strategies}, whereas prior reports on closed-ended benchmarks found clearer gains for more `active' reflection styles~\cite{renze2024SelfReflectionLLMAgents}. The divergence is consistent with an anchor effect: when external signals narrow the solution space, reflection can exploit that signal; when they do not, reflection styles confer little additional benefit over what is afforded from another attempt.

These results are broader than `LLMs are not great at monitoring their own work'. If LLM reflection cannot bind reasoning to specified constraints when external signals are weak or non-existent, reflection will entrench failure modes by rehearsing them, anchoring output on the very material that should be excluded. If scalable intelligence requires dependable self-evaluation, this is a bottleneck; simply adding more scratchpad is unlikely to fix it.

In one study on hallucination by OpenAI \cite{kalai2025WhyLanguageModels}they demonstrated that reinforcement learning can reduce hallucinations on benchmarks but may also encourage responses that are `plausible' or `helpful' rather than an explicit `I don't know' (IDK). Potentially contribution to failures seen on our evaluation. 

Rewarding IDK in RL (as the authors suggest) may help models perform better on our evaluation, since IDK is not a rule violation. However, the issues we observe are not guaranteed to resolve: even if models output IDK instead of a hallucination, their reasoning remains fundamentally incidental and input-contingent. 

For example, in our vignette, the model \emph{mentions the CRT item it plagiarises} as though it were not in the test. It is unclear whether an IDK-trained model would have produced IDK here; but more importantly, even if it did, that would merely prevent plagiarism/rule-violation without revealing the model’s true potential. LLMs clearly possess the knowledge required to respond correctly (e.g., they can list original CRT items when asked); IDK would then be another case where reasoning tokens and outputs are not representative of latent knowledge.

If a more true-to-form `meta-reasoning' were at play, a model would not only gauge certainty but use that signal to `search' for relevant knowledge and apply constraint checks, rather than terminating at IDK. If error detection is itself token-bound, the absence of external signals to `notice' the self-made error and `nudge' toward the right constraint representations will not be solved by IDK alone. Alternatively, calibrated certainty (including IDK) could be used as a control signal inside a reflective loop that triggers retrieval with exclusion filters and constraint verifiers, bringing reflection closer to a mechanism that approximates `meta-reasoning' rather than simulating it.

Lastly, the lack of accessing latent knowledge points to another issue: elicitation of an expert persona did not remedy it. Persona specification is intended to cue the LLM equivalence of human `spreading activation' (as with reflection, we seek only a functional analogue), where input  activate related/nested information to help with reasoning. For example, ask a human accountant `prepare a Business Activity Statement' - relevant constraints such as GST thresholds, exclusion rules, and receipt categories become available in their working memory and constrain what they will file. 

In our experiment, the combination of (i) a CRT-expert persona and (ii) explicit `do not use CRT' instructions failed to elicit constraint fidelity. This speaks to thelimits of persona specificaiton in our open-ended setting.  

Ultimately, our experiments show that evaluation should prioritise open-ended, rule-constrained tests with auditable criteria, not just closed-form unit tests that supply anchors. This surfaces vulnerabilities in reflective reasoning that, if addressed, would improve the reliability of LLM intelligence. Moreover, any development or use of LLM-based systems for automation should prioritise discovering vulnerabilities as in our evaluation, and bind reflection to executable guardrails (constraint verifiers, retrieval with exclusion filters, or human review) until more robust, structural solutions are implemented. Training objectives will need to track error likelihood and rule satisfaction on open-ended tasks if we expect reflection to change outcomes rather than yield confabulatory outputs.

\noindent\textbf{Limitations.} We chose one scientifically meaningful open-ended task—de novo CRT-style item generation with explicit constraints—to maximise auditability. The point is not CRTs per se; it is whether models can apply knowledge under constraints without being handed the answer key. On that requirement, current reflection methods produce persuasive text but inconsistent control. Until reflection binds to constraints, gains will remain modest, variable, and prone to repeating the same mistake in new words.

In addition, our novelty evaluation for `existing non-CRT' item in the `generation' condition may miss low-frequency or newly coined trick-question leading to false negatives, or false positives for example in labelling items as `existing non-crt items' due to superficial similarities. However, our human audits of the labels showed majority agreement in such cases. Our aim was to keep the reflection-loop fully LLM based, however, future evaluations could instead use LLMs with internet access to enhance accuracy.

\bmhead{Acknowledgements}
This work was supported by the Commonwealth Scientific and Industrial Research Organisation (CSIRO) and Aurecon through a doctoral scholarship. In addition, ARC ADM+S has provided support through access to their ADM+S Summer School programme, symposium, and travel support. The funders had no role in study design, data collection and analysis, the decision to publish, or the preparation of the manuscript.

\bmhead{Author contributions}
S.W. conceived the study, designed the task, implemented the code, ran the experiments, performed the analyses, created the figures and tables, and wrote the manuscript. 
F.S. provided supervision throughout, advised study design and analysis strategy, and contributed to interpretation and manuscript editing. 
A.B. provided industry perspective, reviewed the study framing and implications, and contributed comments on the manuscript.

\bmhead{Data and Code Availability}
All code, analysis scripts, and \emph{verbatim prompts} (task templates, reflection templates, evaluator rubric/schema) are available at - https://github.com/cruiseresearchgroup/LLM\_ReflectionTest

\bmhead{Materials \& correspondence}
Correspondence and material requests should be addressed to S.W. (s.weatherhead@unsw.edu.au).

\bmhead{Competing Interests}
The authors declare no competing interests.

\clearpage
\appendix

\section*{Extended Data}
\setcounter{figure}{0}
\setcounter{table}{0}
\renewcommand{\figurename}{Extended Data Fig.}
\renewcommand{\tablename}{Extended Data Table}

\begin{table}[t]
\centering
\small
\begin{tabularx}{\linewidth}{Xrrrr}
\hline
 & \multicolumn{2}{c}{Initial} & \multicolumn{2}{c}{Post-reflection} \\
Model & M & SD & M & SD \\
\hline
claude-3-7-extended      & 0.219 & 0.209 & 0.422 & 0.221 \\
deepseek-reasoner-nvidia & 0.250 & 0.232 & 0.453 & 0.163 \\
gemini-2-5-pro-preview   & 0.281 & 0.088 & 0.516 & 0.087 \\
gpt-4.1                  & 0.188 & 0.222 & 0.383 & 0.270 \\
llama-3-3-70b            & 0.250 & 0.232 & 0.523 & 0.198 \\
llama4-maverick          & 0.125 & 0.189 & 0.305 & 0.321 \\
o3                       & 0.250 & 0.189 & 0.484 & 0.205 \\
o4-mini                  & 0.281 & 0.248 & 0.484 & 0.236 \\
\hline
\end{tabularx}
\caption{Per-model pass-rates (M, SD) at initial generation and after reflection. Values are per-session averages; for reflection, pass-rates are averaged within session across strategies to avoid variance inflation. Omnibus tests found no between-model differences at either round (ANOVA; Kruskal–Wallis robustness checks likewise non-significant).}
\label{ed:model-pass-rates}
\end{table}

\begin{figure}[htbp]
  \centering
  \includegraphics[width=0.88\linewidth]{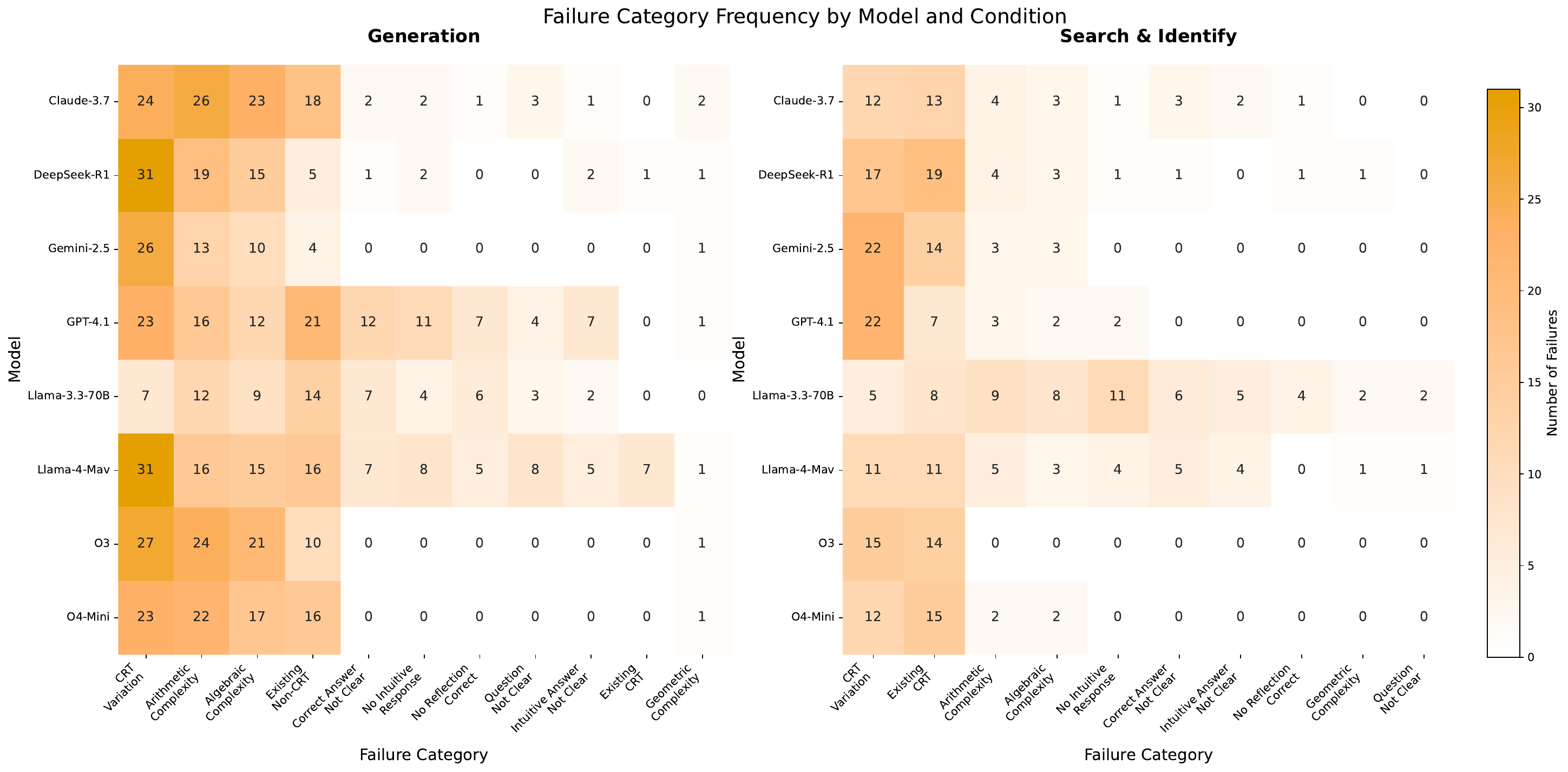}
  \caption{Failure-category frequencies by model and condition. Rows/columns ordered by overall frequency.}
  \label{ed:failcats-heatmap}
\end{figure}

\begin{figure}[htbp]
  \centering
  \includegraphics[width=0.88\linewidth]{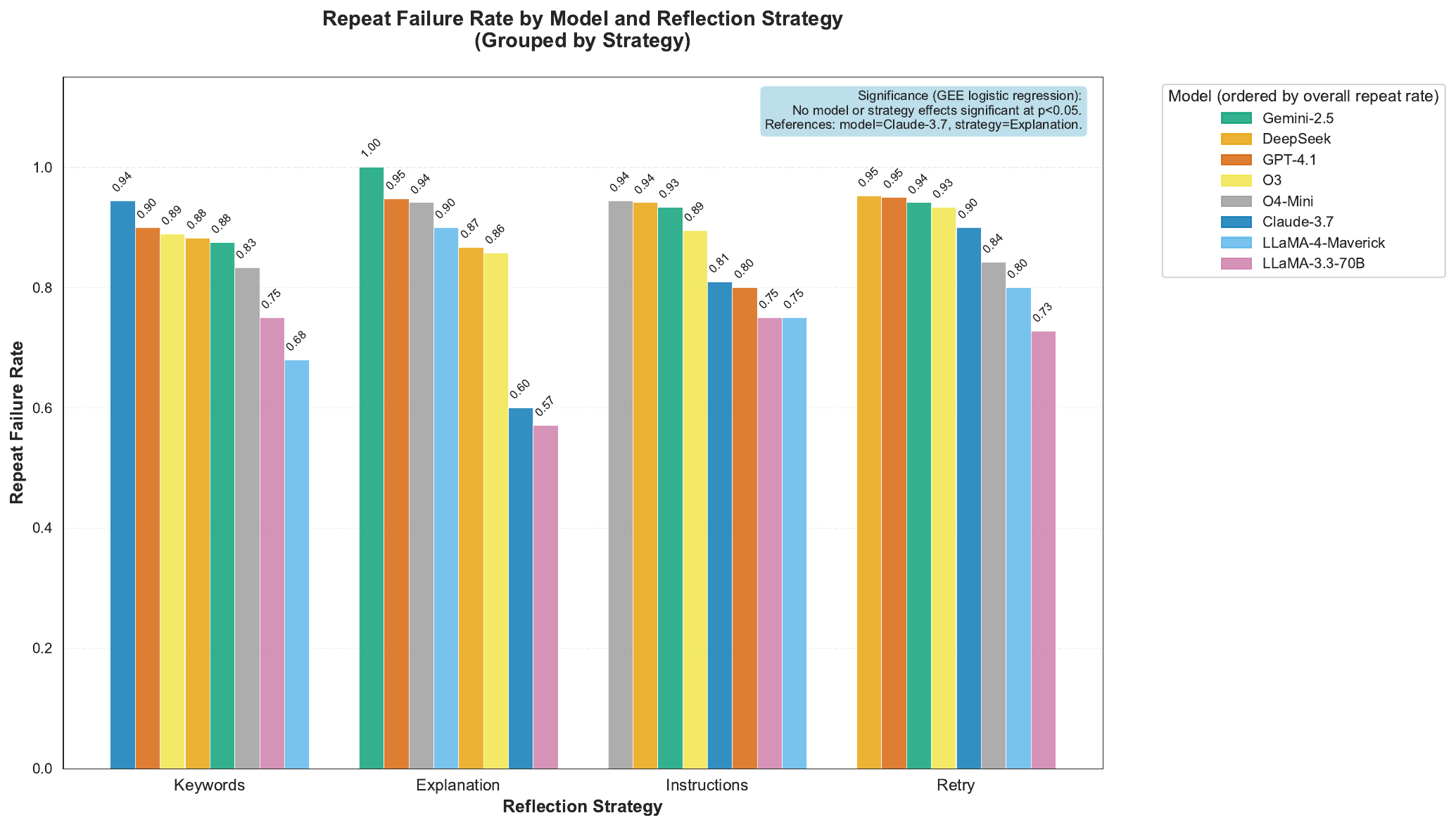}
  \caption{Repeat-failure rate at reflection, grouped by strategy with models colour-coded. No single strategy eliminates repeats across models.}
  \label{ed:repeat-by-strategy-model}
\end{figure}

\begin{figure}[htbp]
  \centering
  \includegraphics[width=0.88\linewidth]{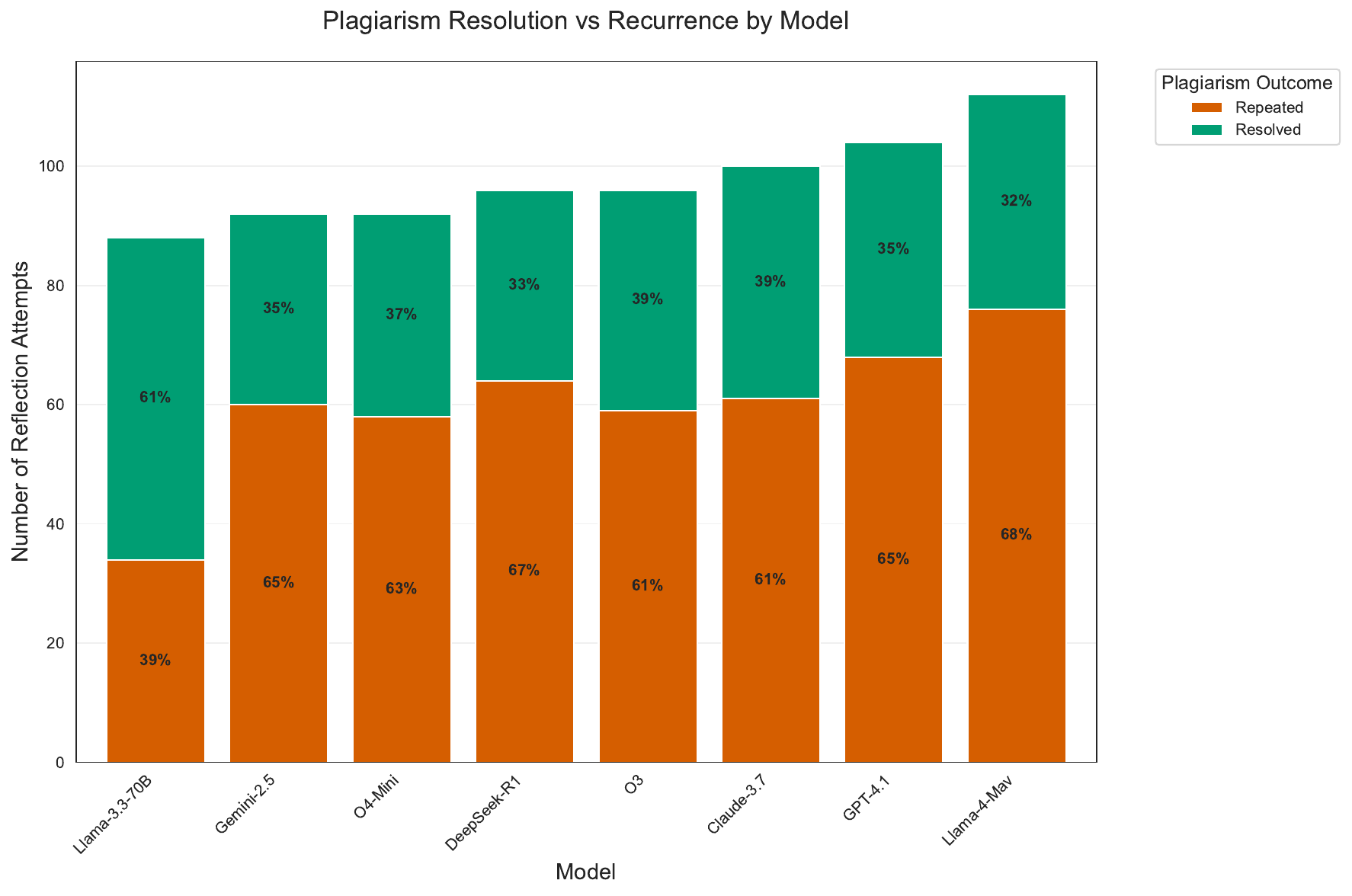}
  \caption{Plagiarism recidivism: share of reflection attempts that plagiarise again, conditional on initial plagiarism. Ordered by repeat proportion.}
  \label{ed:plagiarism-recidivism}
\end{figure}

\begin{figure}[htbp]
  \centering
  \includegraphics[width=0.88\linewidth]{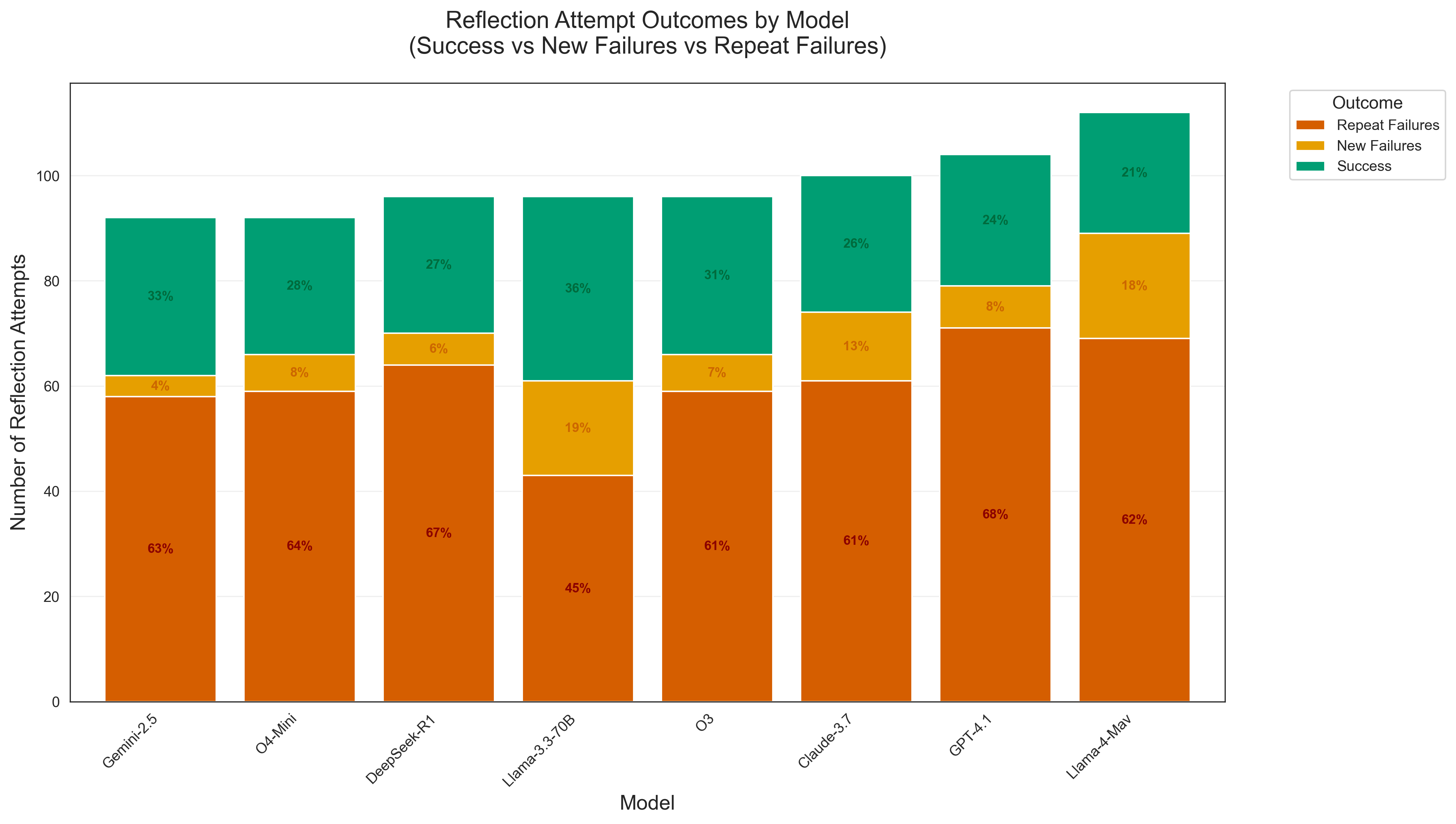}
  \caption{Reflection outcomes: repeat of the original failure category, a new failure category, or corrected (valid).}
  \label{ed:new-failure-outcomes}
\end{figure}

\begin{figure}[htbp]
  \centering
  \includegraphics[width=0.88\linewidth]{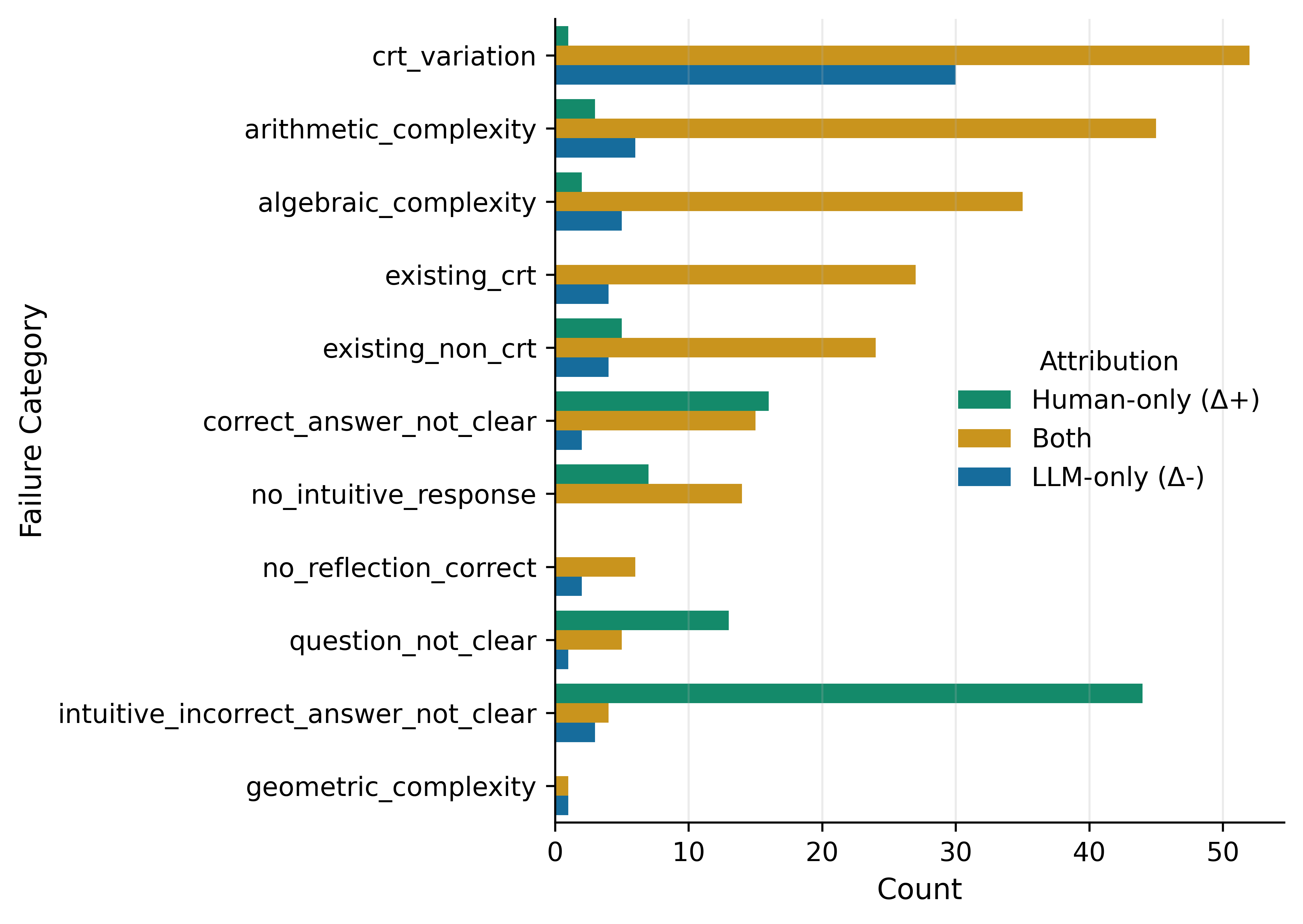}
  \caption{Human–LLM evaluator agreement on category assignments (stratified sample). Agreement is moderate; see supp:human-llm for $\kappa$ values and notes.}
  \label{ed:human-llm-agreement}
\end{figure}

\begin{figure}[htbp]
  \centering
  \includegraphics[width=0.88\linewidth]{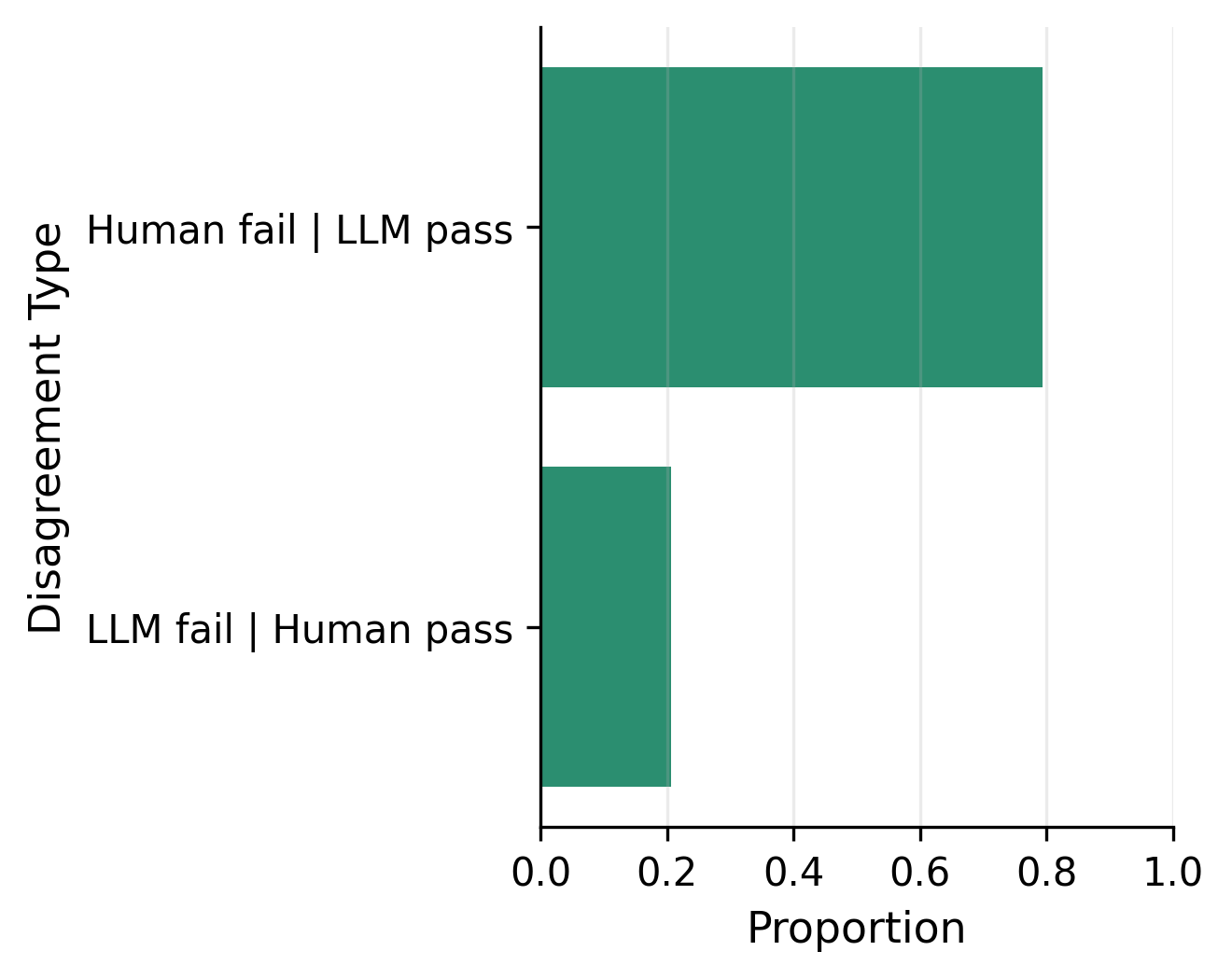}
  \caption{Where humans and LLMs disagree on item fail-pass.}
  \label{ed:human-llm-disagreement}
\end{figure}

\begin{figure}[htbp]
  \centering
  \includegraphics[width=0.88\linewidth]{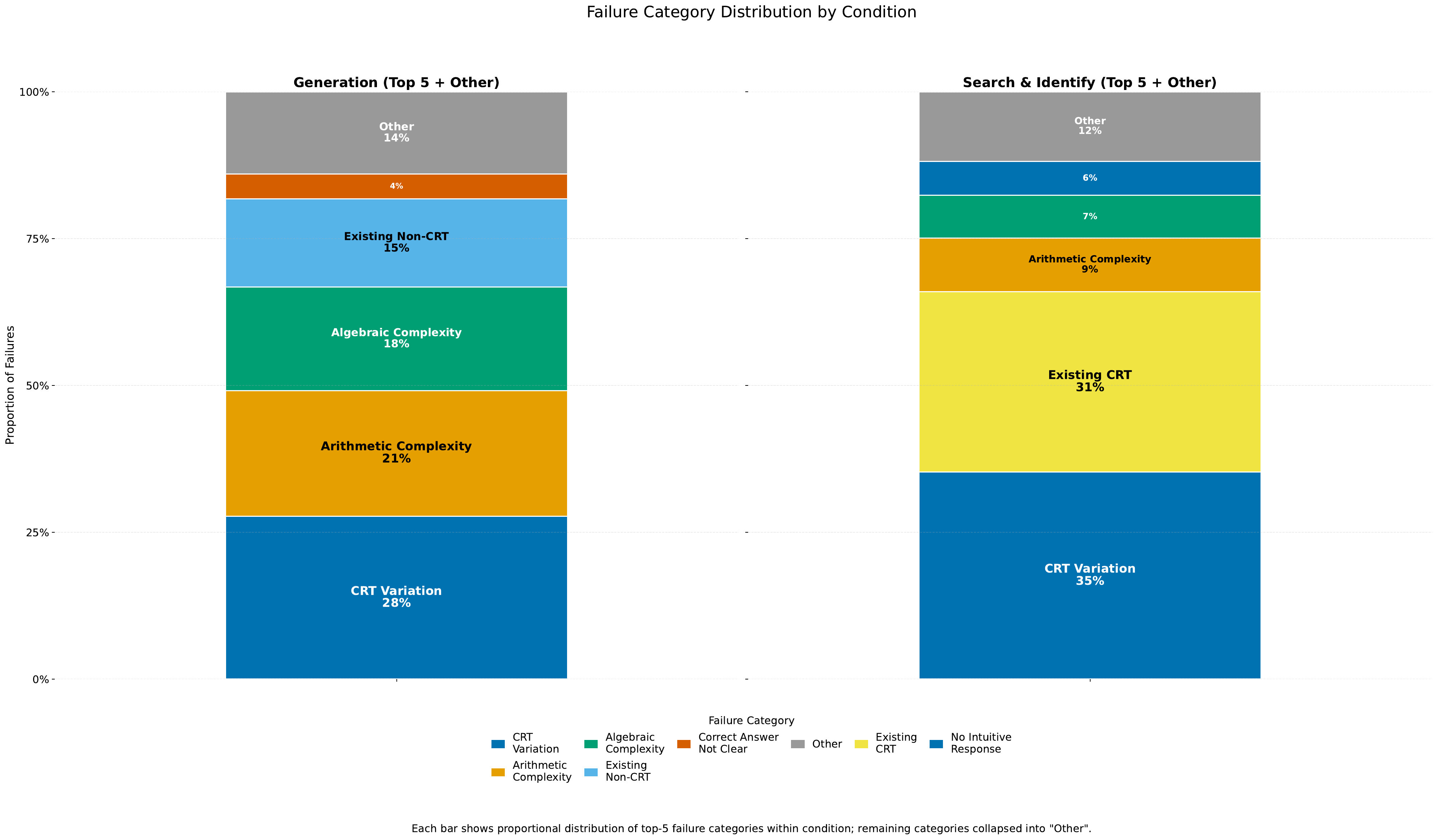}
  \caption{Top-5 failure categories by condition. `Generation' is dominated by \textit{CRT-variation} and arithmetic/algebraic complexity; `search–identify' shifts mass to \textit{existing CRT} and \textit{CRT-variation}.}
  \label{ed:top5-by-condition}
\end{figure}

\begin{figure}[htbp]
  \centering
  \includegraphics[width=0.88\linewidth]{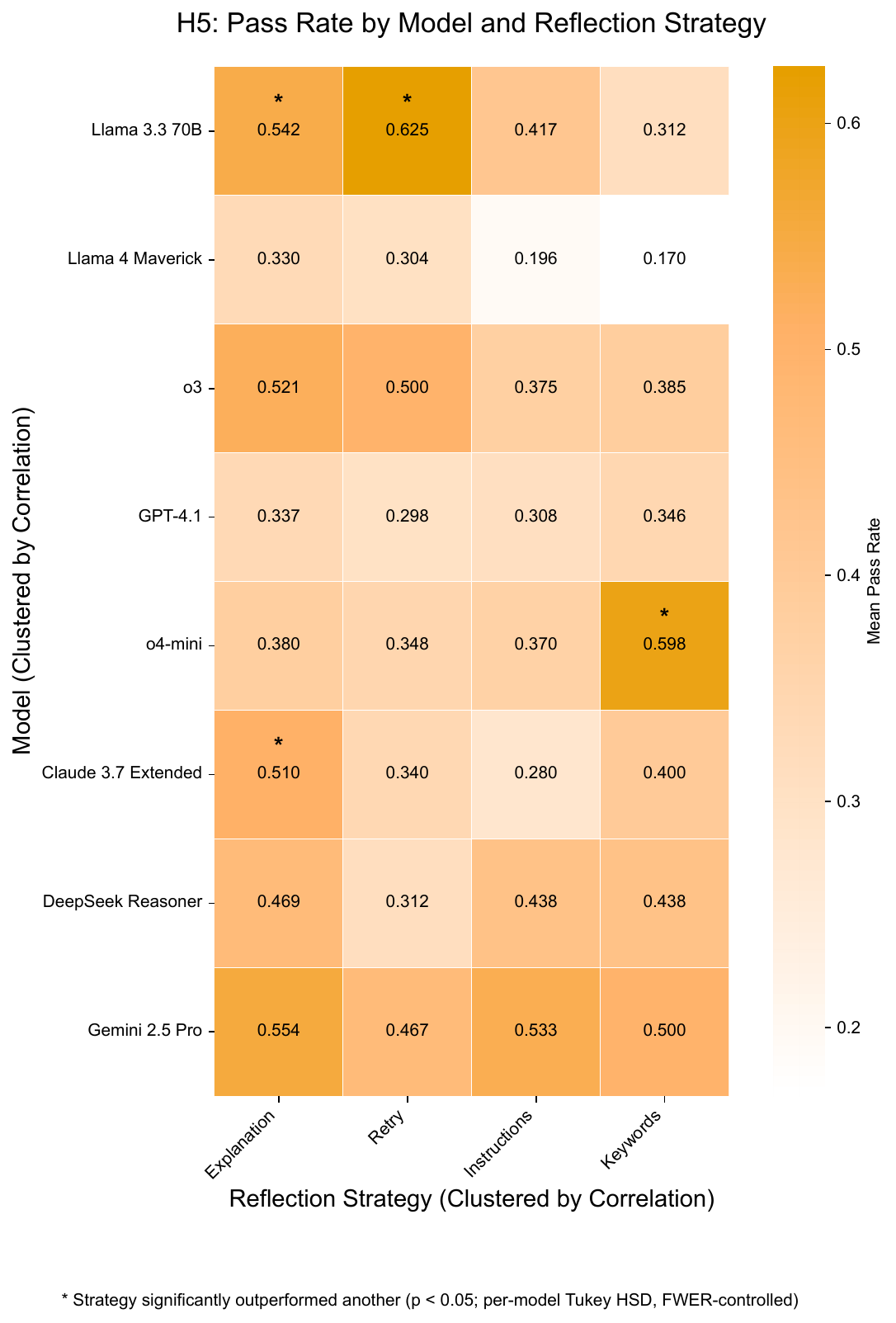}
  \caption{Pass-rates by reflection strategy within model (clustered). `Explanation' leads on average, but gains are model-idiosyncratic; no globally dominant strategy.}
  \label{ed:model-strategy-heatmap}
\end{figure}

\clearpage
\section*{Supplementary Information}

\noindent This Supplementary Information includes further detail for main methods (SA1-SA3) and supplementary analyses we undertook(SA4–SA6). Each block provides a concise Methods note and the corresponding Results, with pointers to Extended Data items where applicable.

\section*{SA1. Robustness checks for H1 (mixed-effects + paired test)}
\paragraph{Methods.}
We estimate a linear mixed–effects model with \textit{generation round} (initial vs.\ reflection) as a fixed effect, plus random intercepts and random slopes by session. As a distribution–free complement, we compute per–session paired differences and report a paired $t$–test with $d_z$.
\paragraph{Results.}
Mixed–effects $\beta_{\mathrm{reflection}}\approx 0.216$ with narrow CIs; paired $t$ corroborates. See Extended Data Table~\ref{ed:model-pass-rates} for per–model descriptives.

\section*{SA2. H1b permutation benchmark for `repeat the original failure'}
\paragraph{Methods.}
Within each task$\times$strategy cell, we permute reflection–round category labels across attempts (10{,}000 draws) to obtain a within–cell chance benchmark (mean and 95\% interval) for `same as initial' repeats. We report observed minus benchmark (percentage points) and a one–sided permutation $p$ for Observed $>$ Benchmark.
\paragraph{Results.}
Observed repeat exceeds the stratified benchmark (see main text H1b). Component category displays are visualised in Extended Data Fig.~\ref{ed:failcats-heatmap} and outcome totals in Extended Data Fig.~\ref{ed:new-failure-outcomes}.

\section*{SA3. Human–LLM evaluator agreement ($\kappa$)}
\paragraph{Methods.}
On a stratified sample we compute human double-code Cohen’s $\kappa$ and Human vs.\ LLM $\kappa$ for category assignment. Disagreements are binned into boundary types (e.g., novelty vs.\ CRT-variation) for illustration.
\paragraph{Results.}
Agreement is moderate, with most disagreements arising at category boundaries. See Extended Data Fig.~\ref{ed:human-llm-agreement}–\ref{ed:human-llm-disagreement}.

\section*{SA4. Strategy effects at reflection (exploratory, per model)}
\paragraph{Methods.}
Reflection–only data with strategy as a fixed effect and a random intercept by session; outcome is $\Delta$ pass-rate relative to the initial attempt. Reference strategy: \textit{Explanation}. Pairwise contrasts use Holm correction; within–model contrasts use Tukey HSD.
\paragraph{Results.}
Strategy gains appear model–idiosyncratic; no global winner. See Extended Data Fig.~\ref{ed:model-strategy-heatmap}.

\section*{SA5. Repeat–failure heterogeneity (logistic GEE)}
\paragraph{Methods.}
Session–clustered logistic GEE for outcome `repeat same category' with fixed effects for model, strategy, and their interaction; marginal effects with 95\% CIs. References: model = Claude-3.7; strategy = \textit{Explanation}.
\paragraph{Results.}
Repeat odds vary modestly by strategy; model effects limited. Extended Data Fig.~\ref{ed:repeat-by-strategy-model} shows the pattern; GEE significance notes are embedded in the caption.

\section*{SA6. Failure prevalence by condition (descriptive)}
\paragraph{Methods.}
We summarise human–coded failure categories by model and condition with bootstrapped CIs; top-$K$ concentration checks are descriptive.
\paragraph{Results.}
`Generation' is dominated by \textit{CRT-variation} and arithmetic/algebraic complexity; `search–identify' shifts mass to \textit{existing CRT} and \textit{CRT-variation}. See Extended Data Fig.~\ref{ed:failcats-heatmap} and Fig.~\ref{ed:top5-by-condition}.

\end{document}